\documentclass[twoside,11pt]{article}

\usepackage{jmlr2e}

\usepackage{tabularx}
\usepackage{booktabs}
\usepackage{float}

\usepackage[acronym]{glossaries}
\newacronym{sdpa}{SDPA}{Scaled Dot Product Attention}
\newacronym{mlp}{MLP}{Multilayer Perceptron}

\usepackage{lastpage}

\firstpageno{1}

\begin{document}

\title{Enhancing compact convolutional transformers with super attention}
\author{
    \name Simpenzwe Honore Leandre\textsuperscript{1,}\textsuperscript{\dag} \email 2101020234@cgu-odisha.ac.in \\
    \addr Dept. of Computer Science and Engineering \\
    C.V. Raman Global University\\
    Bhubaneswar, 752054, Odisha, India
    \AND
    \name Natenaile Asmamaw Shiferaw\textsuperscript{\dag} \email 2101020423@cgu-odisha.ac.in \\
    \addr Dept. of Computer Science and Engineering\\
    C.V. Raman Global University\\
    Bhubaneswar, 752054, Odisha, India
    \AND
    \name Dillip Rout \email dillip.rout.iitb@gmail.com \\
    \addr Dept. of Computer Science and Engineering\\
Assam Royal Global University\\
    Guwahati, 781035, Assam, India
}

\footnotetext[1]{Corresponding author}
\footnotetext[2]{ \dag \hspace{0.05cm}  These authors contributed equally to this work}

\maketitle

\begin{abstract}
In this paper, we propose a vision model that adopts token mixing, sequence-pooling, and convolutional tokenizers to achieve state-of-the-art performance and efficient inference in fixed context-length tasks. In the CIFAR100 benchmark, our model significantly improves the baseline of the top 1\% and top 5\% validation accuracy from 36.50\% to 46.29\% and 66.33\% to 76.31\%, while being more efficient than the \gls{sdpa} transformers when the context length is less than the embedding dimension and only 60\% the size. In addition, the architecture demonstrates high training stability and does not rely on techniques such as data augmentation like mixup, positional embeddings, or learning rate scheduling. We make our code available on \textbf{\underline{\href{https://github.com/Simpenzwe-Honore-Leandre/Compact-vision-transformers-with-super-attention-and-scalable-softmax}{Github}}}.
\end{abstract}


\section{Introduction}

Recent advances like the vision transformer \cite{VIT2020} have become state-of-the-art in the field of computer vision. Since their introduction, significant efforts have been made to improve their efficiency and performance. \cite{CCT2022} introduced the compact convolutional transformer architecture which was able to outperform the class token based vision transformer architecture in data efficiency while being 10x smaller; showing a better scaling potential. However, one drawback of attention is also the prohibitive memory costs it introduces, making it very costly for inference. \cite{super2025} introduced a cost-effective attention mechanisms that could also benefit from optimizations such as Flash Attention, \cite{Flashattention2024} while removing redundant linear transformations.

In this paper, we introduce a variant of compact convolutional transformers that also adopts super attention to improve data efficiency and performance while reducing the number of parameters of the attention layer by 25\% and the total overall parameters by 40\%.

We train a 6-layer model on the CIFAR100 benchmark for 75 epochs and observe an approximately 10\% improvement over \gls{sdpa} in terms of accuracy and faster convergence rate.
\begin{figure}[!htbp]
    \tiny
    \centering
    \includegraphics[width=0.8\textwidth]{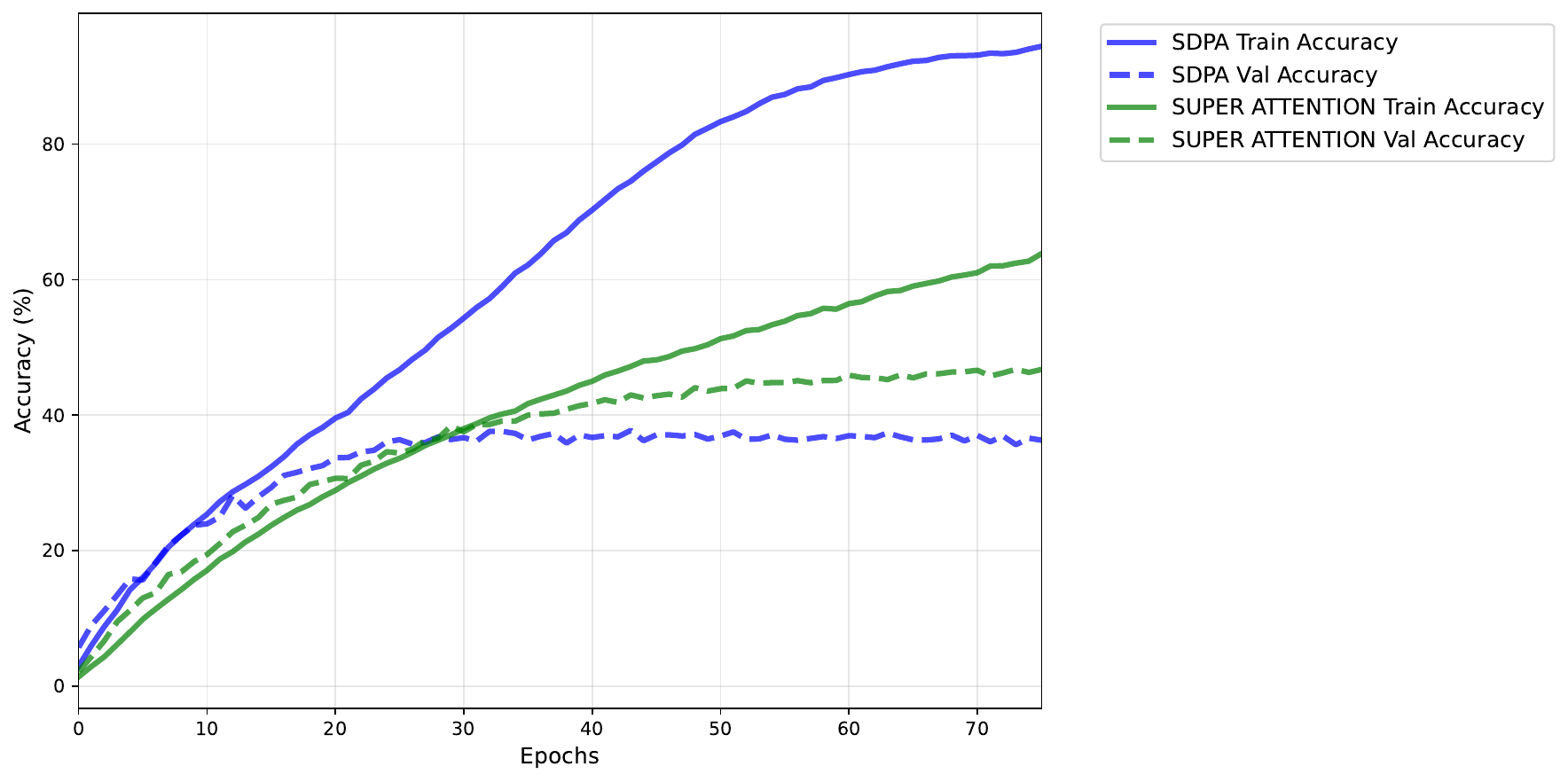}
    \caption{Top 1\% Training and validation accuracy comparison between standard attention (SDPA) and super attention models on CIFAR100 benchmark. Solid lines represent training accuracy, dashed lines show validation accuracy.}
    \label{fig:accuracy_comparison}
\end{figure}

\begin{figure}[!htpb]
    \tiny
    \centering
    \includegraphics[width=0.8\textwidth]{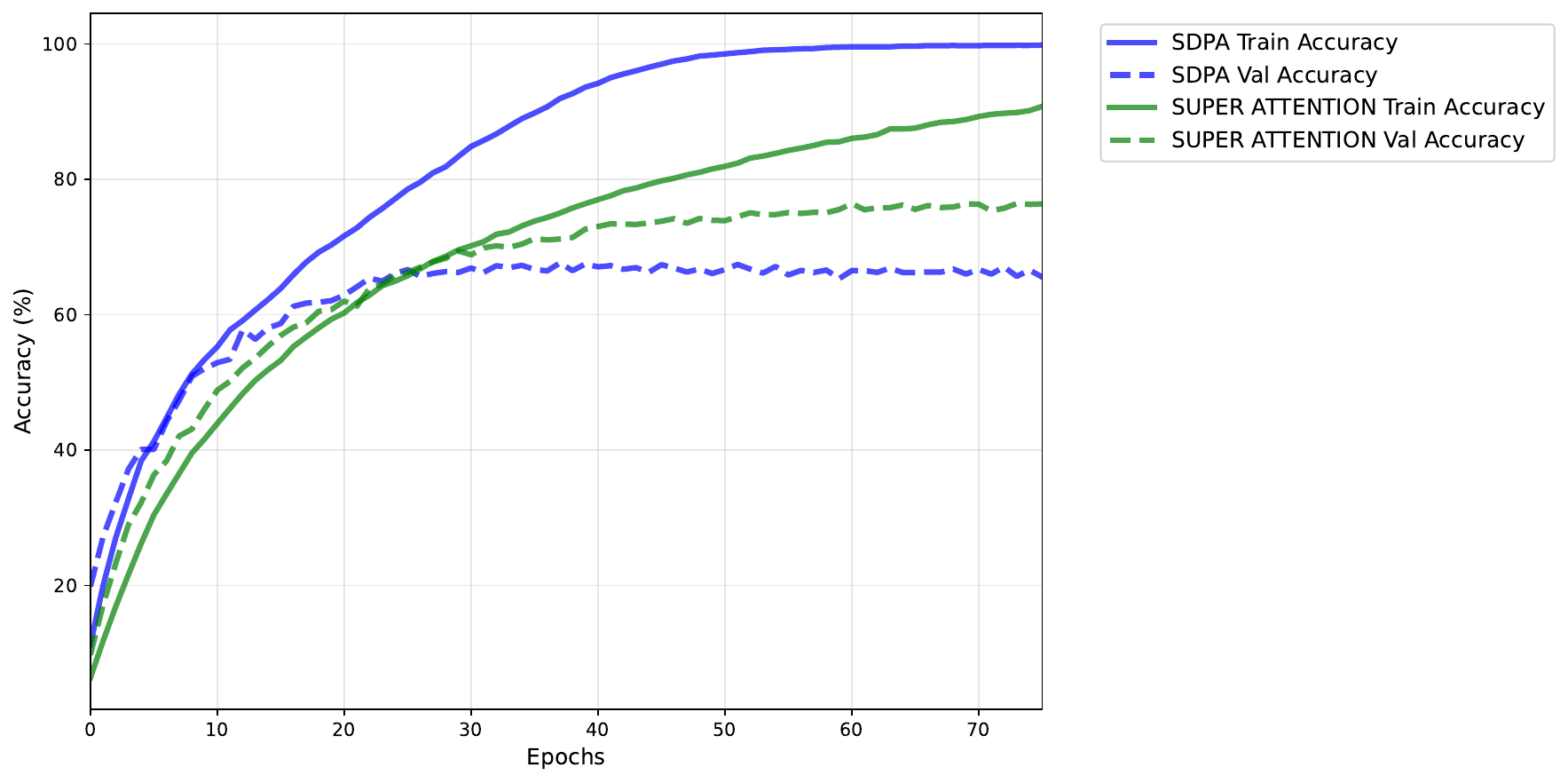}
    \caption{Top 5\% Training and validation accuracy comparison between standard attention (SDPA) and super attention models on CIFAR100 benchmark. Solid lines represent training accuracy, dashed lines show validation accuracy.}
    \label{fig:top 5 accuracy_comparison}
\end{figure}

\section{Literature Review}
\textbf{Optimisation:}
\newline
Building on the original formulation of scaled dot product attention in \cite{vaswani2017}, research such as \cite{Daoflashattention2022} introduced optimization techniques that did not modify the exact formulation to increase training efficiency and reduce memory costs on GPUs. 
Other research such as \cite{super2025} identified redundancies in the original formulation of \gls{sdpa} with respect to the linear transformations that were applied to the q,k and v vectors and introduced token mixing.
\newline
\textbf{Positional embeddings:}
\newline
In \cite{vaswani2017} , positional embeddings were used to learn positional information in the sequence. New approaches like \cite{su2024roformer} were also introduced and have become standard practice. In addition, research such as \cite{haviv2022} showed that causal language models might be able to implicitly learn positional information.
\newline
\textbf{Tokenisation:}
\newline
Popular vision transformer models such as in \cite{VIT2020} adopted patching techniques to tokenize images. The images were split into disjoint patches and flattened to create features. 
In \cite{CVTwu2021}, convolutions were introduced to vision transformers. They were able to safely remove positional encodings while outperforming the vision transformer and ResNet models.
\newline
\textbf{Data augmentation:}
\newline
Data augmentation techniques such as Mixup \cite{zhang2018mixup} and CutMix \cite{yun2019cutmix} are typically used to improve model performance. Mixup generates new training samples by linearly interpolating between pairs of examples, and CutMix extends the idea of Mixup by incorporating spatial information,cutting a rectangular region from one image and pasting it onto another image.

\section{Methodology}

    \subsection{Architecture}
    The transformer block for our architecture and all our experiments is adapted from \cite{deepseekai2025deepseekv3technicalreport}. However, we do not use positional embedding techniques and replace the multihead attention block with super-attention blocks as in Equation \eqref{eq: Super O} then we pool along the sequence length dimension as in Equation \eqref{seqpool}. 

The formulation of token mixing as introduced in \cite{super2025} is defined in the following equations:
  
\begin{center}
\begin{align}
O\ &= (H_1, H_2, \dots, H_h) W^O,\label{eq: Super O}\\
H_i\ &= \ S_i V'_i ,\label{eq: Super H} \\
S_i\ &= softmax(\frac{Q'_i K^T_i}{\sqrt{d_k}}),\label{eq: Super S}\\
V_i'\ &= \ W^A V_i,\label{eq: Super V}\\
Q_i'\ &= \ Q W_i^Q, \label{eq: Super Q}
\end{align}
\end{center}

\begin{itemize}
    \item \textbf{Output}: \(O\) represents (final output)
    \item \textbf{Head-specific terms} (for the \(i\)-th head):
    \begin{itemize}
        \item \(Q'_i\): Query
        \item \(K_i\): Key
        \item \(V'_i\): Value
        \item \(S_i\): Attention score
        \item \(H_i\): Head value
    \end{itemize}
    \item \textbf{Dimensional constants}:
    \begin{itemize}
        \item \(\ell\): Context length
        \item \(d_m\): Embedding dimension/Model dimension
        \item \(h\): Number of heads
        \item \(d_k\): Key dimension (\(W^Q_i, W^K_i \in \mathbb{R}^{d_m \times d_k}\))
        \item \(d_v\): Value dimension (\(W^V_i \in \mathbb{R}^{d_m \times d_v}\))
    \end{itemize}
\end{itemize}

\begin{figure}[H]
    \centering
    \includegraphics[width=0.45\linewidth]{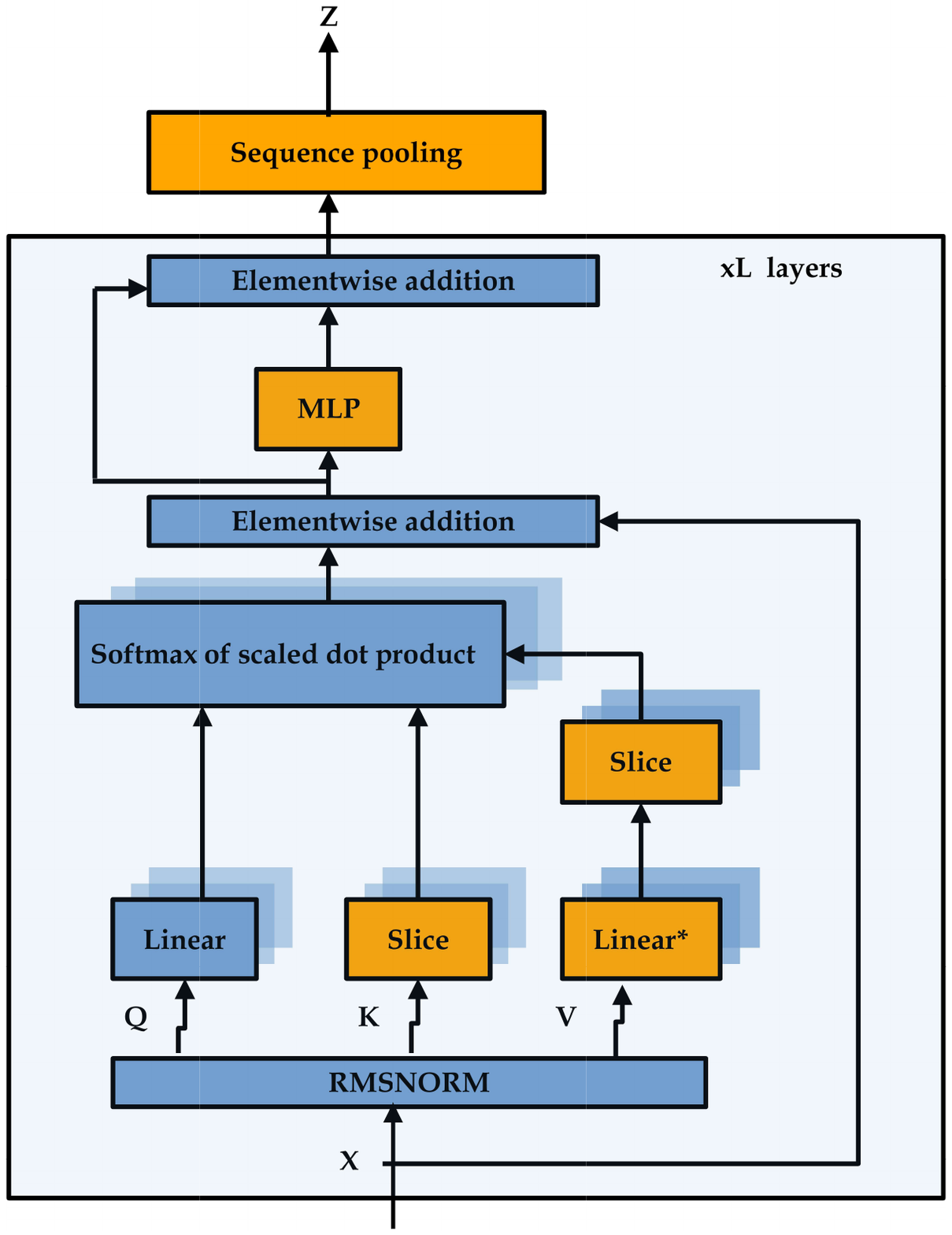}
    \caption{Diagram of transformer backbone. Linear* is transformation from the left.}
    \label{fig:transformer_backbone}
\end{figure}

As seen in equation \eqref{eq: Super H} token mixing is achieved via the \(W^A \in \mathbb{R}^{\ell \times \ell}\) matrix, transforming the values \(V'_i\) from the left.

Architectural advances such as \cite{CCT2022} introduced sequence pooling and convolution-based tokenizers to improve the performance of the vision transformer model.
The sequence pooling mechanism they suggested was defined as:

\begin{center}
    \begin{align}
        O  &= (H_1,H_2,\dots,H_h)W^O \in \mathbb{R}^{b \times \ell \times d_m} \text{ where } b \text{ is the batch size},\\
        O' &= softmax((OW_p)^T) \in \mathbb{R}^{b \times 1 \times \ell} \text{ where } W_p \text{ is a linear transformation } \in \mathbb{R}^{d_m \times 1},\\
        Z &=O'O \in \mathbb{R}^{b \times 1 \times d_m} \label{seqpool}
    \end{align}
\end{center}
The final \(Z\) is then flattened to \(Z \in \mathbb{R}^{b \times d_m}\) and can be sent through a classifier.

\newline
The tokenizer follows \cite{CCT2022} with only one 3x3 convolution layer.
\begin{center}
    \(X'= MaxPool(GELU(Conv2d(X))) \in \mathbb{R}^{b\times \ell \times d_m}\).     
\end{center} 
Where \(X \in \mathbb{R}^{b \times h \times w \times c}\) is an image with b,h,w,c being the batch size, height, width, and number of channels, respectively.
\newline
We also use dropout,  \cite{dropout2014} at a rate of 0.3, an embedding dimension of 768 with 24 attention heads and set all activations to GELU,\cite{GELU2023}.
For equation \eqref{eq: Super O} we use two layer \gls{mlp} instead of a single linear layer.
All model parameters are initialized orthogonally \cite{Orthogonalinit2020}, \cite{saxeorthogonal2014} to speed up convergence.
Our model consists of six transformer layers and a two-layer \gls{mlp} classification head. Figure \ref{fig:transformer_backbone} illustrates the transformer layer architecture.

    \subsection{Dataset}
    For our dataset, we use CIFAR100 due to its diverse class distribution and since it is also a frequently used benchmark to test computer vision models. The CIFAR100 dataset is divided into 100 classes of 600 images each amounting to a total of 60,000 images of size 3x32x32. In total, we have 50000 training images and 10000 validation images.
The images are normalized channel-wise. We do not use any data augmentation techniques.
Due to computation limitations, we do not benchmark other datasets, but expect similar performance regardless.

\section{Experiments}
\begin{table*}[ht]
\centering 
\caption{Parameters Used in the Experiment.} 
\label{table:1}
\begin{tabularx}{\linewidth}{@{}l @{\hskip 3pt} X@{}}
\toprule
\textbf{Parameter Name} & \textbf{Values} \\ 
\midrule
Hyperparameters & Epochs: 75, Batch Size: 1024, Optimizer: AdamW , lr=0.01, \(\beta_1 = 0.9, \beta_2 = 0.999\), weight decay: 0.01\\ 
\midrule
System Specifications & CPU: Intel(R) Xeon(R) CPU @ 2.20GHz, RAM: 12.7 GB, GPU: T4, Storage: 107.7 GB HDD \\ 
\midrule
Operating System & Ubuntu 22.04.4 LTS \\ 
\midrule
Implementation & Python, PyTorch \\ 
\bottomrule
\end{tabularx}
\end{table*}

Our hyperparameter configurations are detailed in Table \ref{table:1}. For simplicity, we maintain a constant learning rate throughout the training.

As shown in Table \ref{tab:accuracy comparison}, our model demonstrates a superior generalization capacity compared to the baseline Compact Convolution Transformer (CCT).

\begin{table*}
\centering
\caption{top-1\% and top-5\% validation accuracy comparison}
\label{tab:accuracy comparison}
\small
\begin{tabularx}{\textwidth}{lXXXXX}
\toprule
\textbf{Model} & \textbf{top 1\%} & \textbf{top 5\%} & \textbf{Params} & \textbf{Epochs}\\ 
\midrule
ViT-12/16 & 39.56\%  & - & 85.63 M & 300\\
CCT-6/3x1 \gls{sdpa} (Our implementation) & 36.50\% & 66.33\% & 17.7 M & 75\\
CCT-6/3x1 Super attention (Ours) & 46.29\%  & 76.31\% & 10.6 M   & 75\\

\bottomrule
\end{tabularx}
\end{table*}

\begin{figure}[H]
    \tiny
    \centering
    \includegraphics[width=0.8\textwidth]{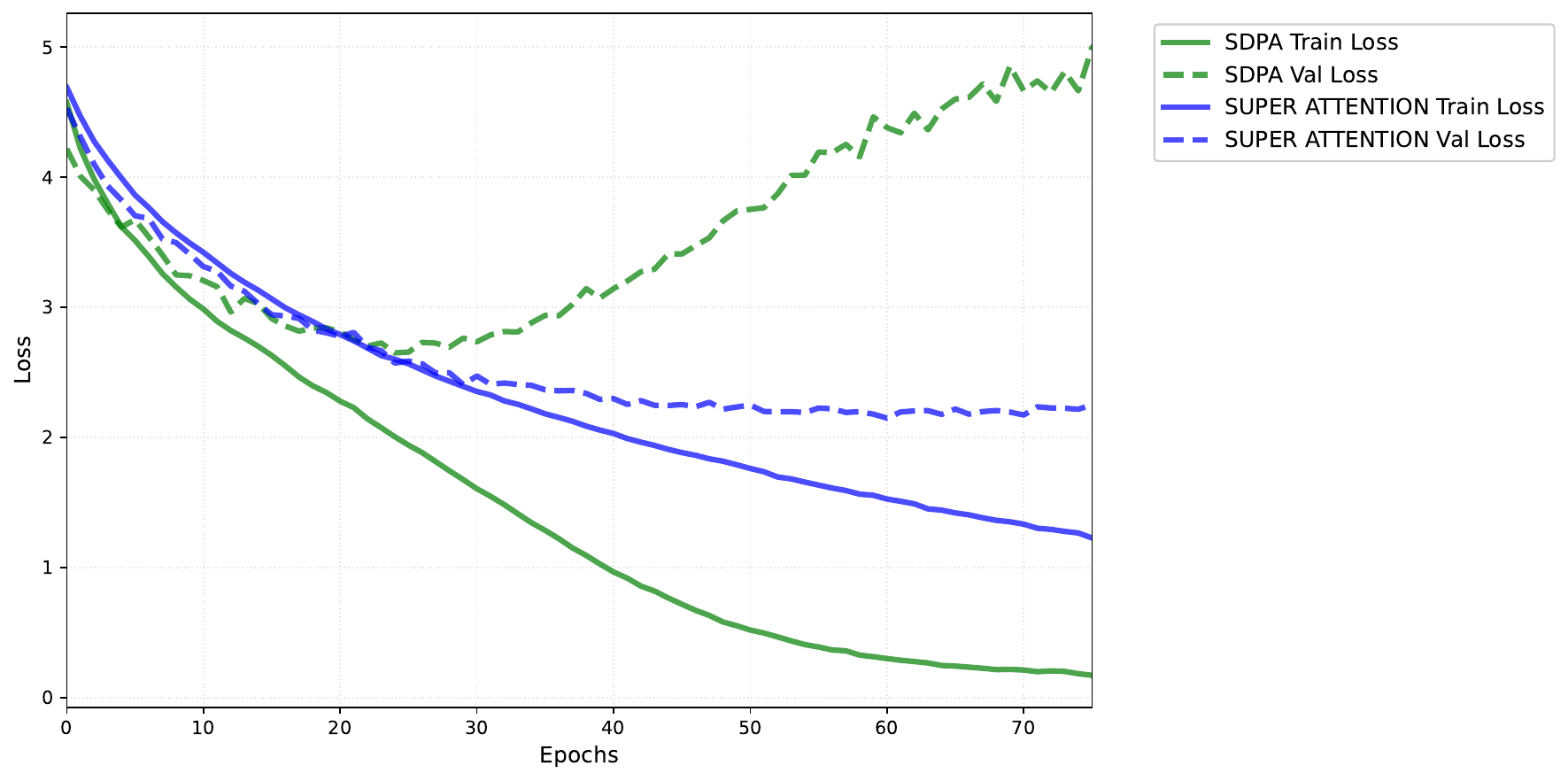}
    \caption{Training and validation loss comparison between standard attention (\gls{sdpa}) and super attention models on CIFAR100 benchmark. Solid lines represent training loss, dashed lines show validation loss.}
    \label{fig:loss_comparison}
\end{figure}

Table \ref{tab:accuracy comparison} shows the accuracy comparison between the baseline and our model.
\newline
For comprehensive benchmarking, we include direct comparisons with ViT-12/16 results from \cite{CCT2022} under identical experimental conditions (specifically, configurations without positional embeddings). This comparison further validates the effectiveness of our approach against established architectures.
\newline
Although the baseline achieves rapid improvements in training accuracy , it exhibits significant overfitting. In contrast, our approach maintains stable convergence. Further analysis in Figure \ref{fig:loss_comparison} reveals that the baseline \gls{sdpa} diverges, whereas our model consistently converges to lower loss values.

\section{Conclusion}
The Vision Transformer remains among the dominant attention-based architectures for visual tasks, benefiting from extensive optimization and widespread adoption. Building on previous advances, particularly the work of \cite{CCT2022} and \cite{super2025}, our approach not only enhances performance but also achieves greater parameter efficiency.\\

This study has several limitations that present opportunities for further research. First, our approach uses a fixed context length, which may constrain performance on generation tasks requiring variable length contexts. Furthermore, computational resource constraints limited the scale of our experiments.

Future research directions could explore:
\begin{itemize}
\item 
Extending the model to handle variable context lengths.
\item 
Incorporating recent optimization techniques like those proposed by \cite{Flashattention2024}.
\item 
Investigating scaling laws through larger model sizes and training runs.
\end{itemize}

These extensions would help to better understand the model's capabilities at scale while improving its practical applicability.

\bibliography{bibliography}

\begin{thebibliography}{16}
\providecommand{\natexlab}[1]{#1}
\providecommand{\url}[1]{\texttt{#1}}
\expandafter\ifx\csname urlstyle\endcsname\relax
  \providecommand{\doi}[1]{doi: #1}\else
  \providecommand{\doi}{doi: \begingroup \urlstyle{rm}\Url}\fi

\bibitem[Dao et~al.(2022)Dao, Fu, Ermon, Rudra, and R\'{e}]{Daoflashattention2022}
Tri Dao, Dan Fu, Stefano Ermon, Atri Rudra, and Christopher R\'{e}.
\newblock Flashattention: Fast and memory-efficient exact attention with io-awareness.
\newblock In S.~Koyejo, S.~Mohamed, A.~Agarwal, D.~Belgrave, K.~Cho, and A.~Oh, editors, \emph{Advances in Neural Information Processing Systems}, volume~35, pages 16344--16359. Curran Associates, Inc., 2022.
\newblock URL \url{https://proceedings.neurips.cc/paper_files/paper/2022/file/67d57c32e20fd0a7a302cb81d36e40d5-Paper-Conference.pdf}.

\bibitem[DeepSeek-AI et~al.(2025)DeepSeek-AI, Liu, Feng, Xue, Wang, Wu, Lu, Zhao, Deng, Zhang, Ruan, Dai, Guo, Yang, Chen, Ji, Li, Lin, Dai, Luo, Hao, Chen, Li, Zhang, Bao, Xu, Wang, Zhang, Ding, Xin, Gao, Li, Qu, Cai, Liang, Guo, Ni, Li, Wang, Chen, Chen, Yuan, Qiu, Li, Song, Dong, Hu, Gao, Guan, Huang, Yu, Wang, Zhang, Xu, Xia, Zhao, Wang, Zhang, Li, Wang, Zhang, Zhang, Tang, Li, Tian, Huang, Wang, Zhang, Wang, Zhu, Chen, Du, Chen, Jin, Ge, Zhang, Pan, Wang, Xu, Zhang, Chen, Li, Lu, Zhou, Chen, Wu, Ye, Ye, Ma, Wang, Zhou, Yu, Zhou, Pan, Wang, Yun, Pei, Sun, Xiao, Zeng, Zhao, An, Liu, Liang, Gao, Yu, Zhang, Li, Jin, Wang, Bi, Liu, Wang, Shen, Chen, Zhang, Chen, Nie, Sun, Wang, Cheng, Liu, Xie, Liu, Yu, Song, Shan, Zhou, Yang, Li, Su, Lin, Li, Wang, Wei, Zhu, Zhang, Xu, Xu, Huang, Li, Zhao, Sun, Li, Wang, Yu, Zheng, Zhang, Shi, Xiong, He, Tang, Piao, Wang, Tan, Ma, Liu, Guo, Wu, Ou, Zhu, Wang, Gong, Zou, He, Zha, Xiong, Ma, Yan, Luo, You, Liu, Zhou, Wu, Ren, Ren, Sha, Fu, Xu, Huang, Zhang, Xie, Zhang, Hao,
  Gou, Ma, Yan, Shao, Xu, Wu, Zhang, Li, Gu, Zhu, Liu, Li, Xie, Song, Gao, and Pan]{deepseekai2025deepseekv3technicalreport}
DeepSeek-AI, Aixin Liu, Bei Feng, Bing Xue, Bingxuan Wang, Bochao Wu, Chengda Lu, Chenggang Zhao, Chengqi Deng, Chenyu Zhang, Chong Ruan, Damai Dai, Daya Guo, Dejian Yang, Deli Chen, Dongjie Ji, Erhang Li, Fangyun Lin, Fucong Dai, Fuli Luo, Guangbo Hao, Guanting Chen, Guowei Li, H.~Zhang, Han Bao, Hanwei Xu, Haocheng Wang, Haowei Zhang, Honghui Ding, Huajian Xin, Huazuo Gao, Hui Li, Hui Qu, J.~L. Cai, Jian Liang, Jianzhong Guo, Jiaqi Ni, Jiashi Li, Jiawei Wang, Jin Chen, Jingchang Chen, Jingyang Yuan, Junjie Qiu, Junlong Li, Junxiao Song, Kai Dong, Kai Hu, Kaige Gao, Kang Guan, Kexin Huang, Kuai Yu, Lean Wang, Lecong Zhang, Lei Xu, Leyi Xia, Liang Zhao, Litong Wang, Liyue Zhang, Meng Li, Miaojun Wang, Mingchuan Zhang, Minghua Zhang, Minghui Tang, Mingming Li, Ning Tian, Panpan Huang, Peiyi Wang, Peng Zhang, Qiancheng Wang, Qihao Zhu, Qinyu Chen, Qiushi Du, R.~J. Chen, R.~L. Jin, Ruiqi Ge, Ruisong Zhang, Ruizhe Pan, Runji Wang, Runxin Xu, Ruoyu Zhang, Ruyi Chen, S.~S. Li, Shanghao Lu, Shangyan Zhou, Shanhuang
  Chen, Shaoqing Wu, Shengfeng Ye, Shengfeng Ye, Shirong Ma, Shiyu Wang, Shuang Zhou, Shuiping Yu, Shunfeng Zhou, Shuting Pan, T.~Wang, Tao Yun, Tian Pei, Tianyu Sun, W.~L. Xiao, Wangding Zeng, Wanjia Zhao, Wei An, Wen Liu, Wenfeng Liang, Wenjun Gao, Wenqin Yu, Wentao Zhang, X.~Q. Li, Xiangyue Jin, Xianzu Wang, Xiao Bi, Xiaodong Liu, Xiaohan Wang, Xiaojin Shen, Xiaokang Chen, Xiaokang Zhang, Xiaosha Chen, Xiaotao Nie, Xiaowen Sun, Xiaoxiang Wang, Xin Cheng, Xin Liu, Xin Xie, Xingchao Liu, Xingkai Yu, Xinnan Song, Xinxia Shan, Xinyi Zhou, Xinyu Yang, Xinyuan Li, Xuecheng Su, Xuheng Lin, Y.~K. Li, Y.~Q. Wang, Y.~X. Wei, Y.~X. Zhu, Yang Zhang, Yanhong Xu, Yanhong Xu, Yanping Huang, Yao Li, Yao Zhao, Yaofeng Sun, Yaohui Li, Yaohui Wang, Yi~Yu, Yi~Zheng, Yichao Zhang, Yifan Shi, Yiliang Xiong, Ying He, Ying Tang, Yishi Piao, Yisong Wang, Yixuan Tan, Yiyang Ma, Yiyuan Liu, Yongqiang Guo, Yu~Wu, Yuan Ou, Yuchen Zhu, Yuduan Wang, Yue Gong, Yuheng Zou, Yujia He, Yukun Zha, Yunfan Xiong, Yunxian Ma, Yuting Yan, Yuxiang
  Luo, Yuxiang You, Yuxuan Liu, Yuyang Zhou, Z.~F. Wu, Z.~Z. Ren, Zehui Ren, Zhangli Sha, Zhe Fu, Zhean Xu, Zhen Huang, Zhen Zhang, Zhenda Xie, Zhengyan Zhang, Zhewen Hao, Zhibin Gou, Zhicheng Ma, Zhigang Yan, Zhihong Shao, Zhipeng Xu, Zhiyu Wu, Zhongyu Zhang, Zhuoshu Li, Zihui Gu, Zijia Zhu, Zijun Liu, Zilin Li, Ziwei Xie, Ziyang Song, Ziyi Gao, and Zizheng Pan.
\newblock Deepseek-v3 technical report, 2025.
\newblock URL \url{https://arxiv.org/abs/2412.19437}.

\bibitem[Dosovitskiy et~al.(2021)Dosovitskiy, Beyer, Kolesnikov, Weissenborn, Zhai, Unterthiner, Dehghani, Minderer, Heigold, Gelly, Uszkoreit, and Houlsby]{VIT2020}
Alexey Dosovitskiy, Lucas Beyer, Alexander Kolesnikov, Dirk Weissenborn, Xiaohua Zhai, Thomas Unterthiner, Mostafa Dehghani, Matthias Minderer, Georg Heigold, Sylvain Gelly, Jakob Uszkoreit, and Neil Houlsby.
\newblock An image is worth 16x16 words: Transformers for image recognition at scale, 2021.
\newblock URL \url{https://arxiv.org/abs/2010.11929}.

\bibitem[Hassani et~al.(2022)Hassani, Walton, Shah, Abuduweili, Li, and Shi]{CCT2022}
Ali Hassani, Steven Walton, Nikhil Shah, Abulikemu Abuduweili, Jiachen Li, and Humphrey Shi.
\newblock Escaping the big data paradigm with compact transformers, 2022.
\newblock URL \url{https://arxiv.org/abs/2104.05704}.

\bibitem[Haviv et~al.(2022)Haviv, Ram, Press, Izsak, and Levy]{haviv2022}
Adi Haviv, Ori Ram, Ofir Press, Peter Izsak, and Omer Levy.
\newblock Transformer language models without positional encodings still learn positional information.
\newblock In Yoav Goldberg, Zornitsa Kozareva, and Yue Zhang, editors, \emph{Findings of the Association for Computational Linguistics: EMNLP 2022}, pages 1382--1390, Abu Dhabi, United Arab Emirates, December 2022. Association for Computational Linguistics.
\newblock \doi{10.18653/v1/2022.findings-emnlp.99}.
\newblock URL \url{https://aclanthology.org/2022.findings-emnlp.99/}.

\bibitem[Hendrycks and Gimpel(2023)]{GELU2023}
Dan Hendrycks and Kevin Gimpel.
\newblock Gaussian error linear units (gelus), 2023.
\newblock URL \url{https://arxiv.org/abs/1606.08415}.

\bibitem[Hosseini et~al.(2025)Hosseini, Hosseini, Castro, and Purver]{super2025}
Peyman Hosseini, Mehran Hosseini, Ignacio Castro, and Matthew Purver.
\newblock Cost-effective attention mechanisms for low resource settings: Necessity \& sufficiency of linear transformations, 2025.
\newblock URL \url{https://arxiv.org/abs/2403.01643}.

\bibitem[Hu et~al.(2020)Hu, Xiao, and Pennington]{Orthogonalinit2020}
Wei Hu, Lechao Xiao, and Jeffrey Pennington.
\newblock Provable benefit of orthogonal initialization in optimizing deep linear networks.
\newblock In \emph{International Conference on Learning Representations}, 2020.
\newblock URL \url{https://openreview.net/forum?id=rkgqN1SYvr}.

\bibitem[Saxe et~al.(2014)Saxe, McClelland, and Ganguli]{saxeorthogonal2014}
Andrew~M. Saxe, James~L. McClelland, and Surya Ganguli.
\newblock Exact solutions to the nonlinear dynamics of learning in deep linear neural networks, 2014.
\newblock URL \url{https://arxiv.org/abs/1312.6120}.

\bibitem[Shah et~al.(2024)Shah, Bikshandi, Zhang, Thakkar, Ramani, and Dao]{Flashattention2024}
Jay Shah, Ganesh Bikshandi, Ying Zhang, Vijay Thakkar, Pradeep Ramani, and Tri Dao.
\newblock Flashattention-3: Fast and accurate attention with asynchrony and low-precision, 2024.
\newblock URL \url{https://arxiv.org/abs/2407.08608}.

\bibitem[Srivastava et~al.(2014)Srivastava, Hinton, Krizhevsky, Sutskever, and Salakhutdinov]{dropout2014}
Nitish Srivastava, Geoffrey Hinton, Alex Krizhevsky, Ilya Sutskever, and Ruslan Salakhutdinov.
\newblock Dropout: A simple way to prevent neural networks from overfitting.
\newblock \emph{Journal of Machine Learning Research}, 15\penalty0 (56):\penalty0 1929--1958, 2014.
\newblock URL \url{http://jmlr.org/papers/v15/srivastava14a.html}.

\bibitem[Su et~al.(2024)Su, Ahmed, Lu, Pan, Bo, and Liu]{su2024roformer}
Jianlin Su, Murtadha Ahmed, Yu~Lu, Shengfeng Pan, Wen Bo, and Yunfeng Liu.
\newblock Roformer: Enhanced transformer with rotary position embedding.
\newblock \emph{Neurocomputing}, 568:\penalty0 127063, 2024.
\newblock ISSN 0925-2312.
\newblock \doi{https://doi.org/10.1016/j.neucom.2023.127063}.
\newblock URL \url{https://www.sciencedirect.com/science/article/pii/S0925231223011864}.

\bibitem[Vaswani et~al.(2017)Vaswani, Shazeer, Parmar, Uszkoreit, Jones, Gomez, Kaiser, and Polosukhin]{vaswani2017}
Ashish Vaswani, Noam Shazeer, Niki Parmar, Jakob Uszkoreit, Llion Jones, Aidan~N Gomez, \L~ukasz Kaiser, and Illia Polosukhin.
\newblock Attention is all you need.
\newblock In I.~Guyon, U.~Von Luxburg, S.~Bengio, H.~Wallach, R.~Fergus, S.~Vishwanathan, and R.~Garnett, editors, \emph{Advances in Neural Information Processing Systems}, volume~30. Curran Associates, Inc., 2017.
\newblock URL \url{https://proceedings.neurips.cc/paper_files/paper/2017/file/3f5ee243547dee91fbd053c1c4a845aa-Paper.pdf}.

\bibitem[Wu et~al.(2021)Wu, Xiao, Codella, Liu, Dai, Yuan, and Zhang]{CVTwu2021}
Haiping Wu, Bin Xiao, Noel Codella, Mengchen Liu, Xiyang Dai, Lu~Yuan, and Lei Zhang.
\newblock Cvt: Introducing convolutions to vision transformers, 2021.
\newblock URL \url{https://arxiv.org/abs/2103.15808}.

\bibitem[Yun et~al.(2019)Yun, Han, Oh, Chun, Choe, and Yoo]{yun2019cutmix}
Sangdoo Yun, Dongyoon Han, Seong~Joon Oh, Sanghyuk Chun, Junsuk Choe, and Youngjoon Yoo.
\newblock Cutmix: Regularization strategy to train strong classifiers with localizable features, 2019.
\newblock URL \url{https://arxiv.org/abs/1905.04899}.

\bibitem[Zhang et~al.(2018)Zhang, Cisse, Dauphin, and Lopez-Paz]{zhang2018mixup}
Hongyi Zhang, Moustapha Cisse, Yann~N. Dauphin, and David Lopez-Paz.
\newblock mixup: Beyond empirical risk minimization, 2018.
\newblock URL \url{https://arxiv.org/abs/1710.09412}.

\end{thebibliography}

\end{document}